\documentclass{article}
\usepackage{spconf,amsmath,graphicx}
\usepackage{pifont,booktabs,amssymb,multirow,hyperref}

\title{an empirical study and improvement for speech emotion recognition}
\name{Zhen Wu\thanks{Zhen Wu and Yizhe Lu contributed equally. This work was supported by the National Natural Science Foundation of China (No. 62206126 and 61976114). Our code is available at \url{https://github.com/Luyizhe/SpeechEmotion}.}, Yizhe Lu, Xinyu Dai}
\address{National Key Laboratory for Novel Software Technology, Nanjing University\\
	Collaborative Innovation Center of Novel Software Technology and Industrialization, Nanjing\\
	\texttt{wuz@nju.edu.cn, luyz@smail.nju.edu.cn, daixinyu@nju.edu.cn}}
\begin{document}
%
\maketitle
\begin{abstract}
Multimodal speech emotion recognition aims to detect speakers' emotions from audio and text. Prior works mainly focus on exploiting advanced networks to model and fuse different modality information to facilitate performance, while neglecting the effect of different fusion strategies on emotion recognition. In this work, we consider a simple yet important problem: \emph{how to fuse audio and text modality information is more helpful for this multimodal task}. Further, we propose a multimodal emotion recognition model improved by perspective loss. Empirical results show our method obtained new state-of-the-art results on the IEMOCAP dataset. The in-depth analysis explains why the improved model can achieve improvements and outperforms baselines. 
\end{abstract}
\section{Introduction}
\label{sec:intro}

Speech emotion recognition is a crucial subtask of affective computing, which aims to identify the speaker's emotions from user utterances~\cite{poria2017multi}. In recent years, this task has drawn the increasing attention of researchers and industries because of its wide applications in social media analysis, human-computer interaction, and  customer service.

Early works use audio-only information to identify speech emotion~\cite{mirsamadi2017automatic,chiba2020multi,gao2021domain}. For example, Mirsamad et al.~\cite{mirsamadi2017automatic} employ attention-based RNN to capture emotion-aware features. Gao et al.~\cite{gao2021domain} eliminate the influence of different speakers in the audio by adversarial neural networks. In fact, speech naturally contains multimodal audio and text information, and they embody emotions from phonetics and linguistics respectively. Realizing this characteristic, the following researchers use multimodal information to build neural models. These works~\cite{krishna2020multimodal,xu2019learning,Yoonhyung2020Multimodal,yoon2019speech,yoon2020attentive} widely use cross-attention to fuse information from different modalities and achieve a considerable improvement over unimodal methods in accuracy. 

The above studies all focus on the emotional recognition of a single utterance. Considering the relevance between utterances, more recent works step towards the actual scenario, conversational multimodal emotion recognition. Poria et al.~\cite{poria2017context} adopt hierarchical modeling, and they extract utterance-level features first and then aggregate dialogue-level information. In the following work~\cite{poria2017multi}, they propose an attention-based fusion (AT-Fusion) approach~\cite{poria2017multi} to obtain different modal utterance representations and get the global information by additional LSTM~\cite{hochreiter1997long} and self-attention mechanism~\cite{Ashish2017Attention}. Besides, graph neural networks are also employed to fuse audio and text modality features~\cite{hu2022mm,fu2021consk}.

Despite the effectiveness, the existing works rely on advanced neural networks to achieve remarkable results. They neglect the effect of fusion manners of different modalities on emotion recognition. In this work, we revisit the issue and answer one crucial question: \emph{what fusion manner is more effective for multimodal emotion recognition model?}

To achieve the goal, we extract a general unimodal audio/text framework from existing research as a fundamental component and investigate the effect of various ways of fusing multimodal information at different layers. We find some simple operations have achieved very competitive performance. Further, we proposed an improved multimodal model with the perspective loss and achieved new state-of-the-art results on the IEMOCAP~\cite{busso2008iemocap} dataset. Empirical results and analysis explain why the multimodal model and our improved version can achieve better performance.

\section{APPROACH}
\label{sec:appr}
We define the emotion recognition task in conversation first and then show different fusion methods. Finally, we introduce our perspective loss. 

\subsection{Problem Definition}
\noindent A set of dialogue $\mathcal{U}=\{U_1, U_2, . . . , U_N\} $ contains $N$ dialogues. Each dialogue $U=\{(u_1, e_1), (u_2, e_2), ... , (u_S, $ $e_S)\}$ consists of $S$ utterances and $u_i$ is its $i_{th}$ utterance. The emotion label of utterance $u_i$ is $e_i\in$\{\emph{happy}, \emph{sad}, \emph{neutral}, \emph{angry}\}. The task is to build a model to fit the $N$ labeled dialogues $\mathcal{U}$ and then predict the emotion labels of dialogues in the test set. 

\subsection{Unimodal Framework}
\label{unimodalframe}
\noindent Before applying different fusion methods, we introduce our unimodal audio/text framework.

\begin{figure}
\centering
\includegraphics[width=50mm]{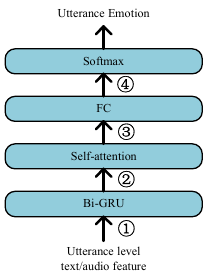}
\caption{Overview of unimodal structure.}
\label{fig1:env}
\end{figure}

As illustrated in Fig.~\ref{fig1:env}, the general unimodal framework is extracted from the existing multimodal framework~\cite{poria2017multi,poria2017context,jiao2019higru}. First, we extract utterance-level audio features or text features (see section~\ref{sec:features}). For the unimodal features, we then apply a bi-directional GRU (Bi-GRU) to encode all utterances and capture the past and future information of the dialogue. Next, we use the outputs of the Bi-GRU layer as the inputs of the self-attention module~\cite{poria2017multi}, which gets rich contextual information from relevant utterances. After that, the outputs of the self-attention module are sent to a fully-connected (FC) layer, which maps high-level features to dimensions of emotion categories. \textbf{Note that}, the FC's output is also called logits. Finally, a softmax layer can be applied to transform logits into probabilities for unimodal emotion classification. 

\subsection{Fusion Methods}
\noindent Unimodal framework contains multiple stacked components. To analyze the effect of multimodal fusion at different layers, we explore four positions:  \ding{172} Early Fusion (EF), \ding{173} Middle Fusion (MF), \ding{174} Late Fusion (LF), and \ding{175} Logits Fusion (LGF), as shown in Fig. \ref{fig1:env}.

In terms of fusion strategy, there are three general methods of multimodal fusion in the literature, respectively ADD-Fusion, Concat-Fusion, and AT-Fusion. To describe them clearly, we take Logits Fusion (LGF) as an example and introduce some necessary notations. Specifically, $a_i$ $\in$ $\mathbb{R}^{d}$ and $t_i$ $\in$ $\mathbb{R}^{d}$ respectively represent the audio and text representations from unimodal position \ding{175} before fusion. Here audio representation and text representation have the same dimension of $d$ because we map original audio and text features to the same dimension early when extracting multimodal features (section~\ref{sec:features}).

\textbf{ADD-Fusion.} In this manner, we directly add two different modal features to obtain the fusion representation $z_i$ $\in$ $\mathbb{R}^{d}$. The process is as follows:
\begin{equation}
    z_i=a_i+ t_i.
\end{equation}

\textbf{Concat-Fusion.} In this strategy, different modal features are concatenated together and then mapped to the target dimension. The fused representation $z_i$ $\in$ $\mathbb{R}^{d}$ is as follows:
\begin{equation}
    z_i=W_f\cdot[a_i; t_i],
\end{equation}
where $W_f$ $\in$ $\mathbb{R}^{d \times 2d}$ and $[;]$ denotes concatenation. 

\textbf{AT-Fusion.} AT-Fusion fuses different modal features in the form of a weighted sum, where weights measure modality importance. The detailed calculation process is the same as the fusion process of \cite{poria2017multi}. 

\subsection{Perspective Loss}
\noindent Usually, the fused representation $z_i$ can be used for emotion classification directly. The corresponding cross-entropy loss is marked as \textbf{Single Loss}:
\begin{equation}
    \mathcal{L}_s=\sum^{N}_{U\in\mathcal{U}}\sum^{S}_{i=1}-\text{log}P(\hat{y}_{i}=y_{i}),
\end{equation}
where $\hat{y}_i$ and $y_{i}$ are respectively the predicted label and truth label of utterance $u_i$ in the dialogue $U$.

It is common sense that audio and text modalities contain diverse features, e.g., phonetics and linguistics. Inspired by the work~\cite{wu2018improving},
we introduced additional cross-entropy losses of emotion classification respectively for audio and text modalities to encourage the model to retain helpful unimodal characteristics for emotion when fusing multimodal information. Finally, we combined these two unimodal losses and the above \textbf{Single Loss} to form \textbf{Perspective Loss}:
\begin{align}
    \mathcal{L}_a&=\sum^{N}_{U\in\mathcal{U}}\sum^{S}_{i=1}-\text{log}P(\hat{y}^{a}_{i}=y_{i}),\\
    \mathcal{L}_t&=\sum^{N}_{U\in\mathcal{U}}\sum^{S}_{i=1}-\text{log}P(\hat{y}^{t}_{i}=y_{i}),\\
    \mathcal{L}_p&=\mathcal{L}_a+\mathcal{L}_t+\mathcal{L}_s,
\end{align}
where $\hat{y}^{a}_i$ and $\hat{y}^{t}_i$ respectively represent the predicted labels of audio modality and text modality.

\section{EXPERIMENTAL SETTINGS}
\label{sec:exper}
\subsection{Dataset}
\noindent We conduct experiments on the IEMOCAP~\cite{busso2008iemocap} dataset. It is produced by 10 speakers in $5$ sessions and contains hours of various dialogue scenarios. Following previous works~\cite{poria2017context,dutta2022multimodal}, we use four categories where \emph{happy} and \emph{excited} are merged into the \emph{happy} category. So the final dataset contains 5531 utterances (\emph{happy}: 1636, \emph{neutral}: 1084, \emph{angry}: 1103, \emph{sad}: 1708). In the experiments, we set $1-4$ sessions as the training set and the $5$th session as the test set.
\subsection{Features Extraction}
\label{sec:features}
\noindent In this work, we extract the unimodal audio and text utterance features with different tools.
\subsubsection{Audio Features} 
Following the work~\cite{hazarika2018conversational}, we use the software openSMILE with the config file \textit{IS13\_ComParE} to extract 6373 features for each utterance. Z-standardization is performed on features for normalization. We then employ an FC layer to reduce the dimension of the audio vector to 100, which is used as the final utterance-level features of audio.

\subsubsection{Text Features} 
For each utterance, we use the pre-trained $\text{BERT}_{base}$ to encode text and adopt mean pooling of all token representations to obtain a vector with 768 dimensions. After that, we use an FC layer to reduce the vector to 100 dimensions as the utterance features of text modality.  

\begin{table}[t]
\begin{center}
\setlength{\tabcolsep}{7mm}{
\begin{tabular}{c c }
    \toprule[1pt]
    \textbf{Modality}    & \textbf{WA(\%)} \\
    \hline
    Audio   & 71.19 \\
    Text    & 82.58 \\
    \bottomrule[1pt]
\end{tabular}}
\caption{The WA(\%) results of audio unimodal framework and text unimodal framework mentioned in section~\ref{unimodalframe} on speech emotion classification.}
\label{table:tab1}
\end{center}
\end{table}

\subsection{Hyper-parameters and Metric}
\noindent We set the hidden states of Bi-GRU network to 100 dimensions. The self-attention layer contains 100-dimensional states and 4 attention heads. We use ReLU as the nonlinear activation layer. The Adam optimization is adopted to optimize the model with a learning rate of 0.001 and decay of 0.00001. We train the models for 150 epochs with a batch size of 20, dropout with $p=0.2$, and $L2$ regularization with the weight of 0.00001. To obtain statistically credible results, we repeat the experiments 20 times with different initial weights and report the average result to reduce the randomness. Weighted accuracy (WA)~\cite{mirsamadi2017automatic} is chosen as our evaluation metric to compare our method with previous approaches. 

\section{Results and analysis}
\label{sec:ana}
\noindent In this section, we show the effects of multimodality, different fusions, and perspective loss on speech emotion recognition. Then we compare the results of the proposed model with other state-of-the-art methods.

\begin{table}[t]
\begin{center}
\small
\setlength{\tabcolsep}{2mm}{
\begin{tabular}{c c c c}\toprule[1pt]
    \textbf{FL}    & \textbf{FM} & \textbf{WA(\%) $\mathcal{L}_s$}
    & \textbf{WA(\%) $\mathcal{L}_p$} 
    \\
    \hline
                    & ADD-Fusion    & {83.95$\pm$0.64} & {84.57$\pm$0.52} \\
    EF    & Concat-Fusion & {83.12$\pm$0.85} & {83.68$\pm$0.52} \\
                    & AT-Fusion     & {83.12$\pm$0.59} & {83.29$\pm$0.51} \\
    \hline
                    & ADD-Fusion    & {83.86$\pm$0.54} & {85.23$\pm$0.54} \\
    MF    & Concat-Fusion & {83.25$\pm$0.79} & {84.89$\pm$0.34} \\
                    & AT-Fusion     & {83.12$\pm$1.01} & {84.99$\pm$0.22} \\
    \hline
                    & ADD-Fusion    & {83.61$\pm$0.46} & {84.80$\pm$0.41} \\
    LF    & Concat-Fusion & {83.54$\pm$1.77} & {84.43$\pm$0.55} \\
                    & AT-Fusion     & {83.20$\pm$1.72} & {84.77$\pm$0.63} \\
    \hline
                    & ADD-Fusion    & {83.39$\pm$0.54} & {\textbf{85.40}$\pm$0.40} \\
    LGF    & Concat-Fusion & {83.23$\pm$1.57} & {84.60$\pm$0.47} \\
                    & AT-Fusion     & {81.79$\pm$1.37} & {84.52$\pm$0.72} \\
\bottomrule[1pt]
\end{tabular}}
\caption{The WA(\%) results of multimodal models with different fusion methods and layers on the IEMOCAP dataset. FL means Fusion Layer, and FM is Fusion Method. WA(\%) $\mathcal{L}_s$ denotes using single loss $\mathcal{L}_s$ to train models, while WA(\%) $\mathcal{L}_p$ is to train models with perspective loss $\mathcal{L}_p$. The numbers after ``$\pm$'' represent standard deviations.}
\label{table:tab2}
\end{center}
\end{table}

\subsection{Unimodal v.s. Multimodal}
Table~\ref{table:tab1} with Table~\ref{table:tab2} respectively show the results of our unimodal framework and multimodal framework on the IEMOCAP dataset. As in Table~\ref{table:tab1}, the performance of text modality surpasses audio modality significantly. It is reasonable because we observe most speech emotions can be inferred from texts directly. After fusing audio and text modal information, multimodal models outperform unimodal models in most settings. This phenomenon shows that the information in the two modalities can be complementary. It provides the possibility of designing various multimodal fusion models to improve speech emotion recognition.

\subsection{Effects of Different Fusions}
As shown in Table~\ref{table:tab2}, regarding different fusion methods, we find ADD-Fusion performs the best in the three methods wherever we conduct fusion operations. This may be attributed to that ADD-Fusion operation  utilizes two modalities equally and make them complementary better in feature space. In contrast, Concat-Fusion and AT-Fusion may lose focus on certain modalities due to the weighted combination. 

When fusing at different layers, EF with ADD-Fusion shows better performance if using Single Loss, which indicates early fusion makes multimodal information interact more fully. In contrast, LGF performs the best when we use Perspective Loss to enhance modality characteristics. We guess Perspective Loss makes the features in LGF decoupled and contributes to emotion recognition from different views. 

\begin{table}[t]
\begin{center}
\small 
\setlength{\tabcolsep}{1mm}{
\begin{tabular}{c | c | c c c c c}\toprule[1pt]
     & &\multicolumn{4}{c}{Emotion on IEMOCAP(\%)}\\
    \hline
    {Model}                       & {Modality}  & {happy}   & {sad} & {neutral} & {angry}  \\
    \hline
    \multirow{2}{*}{Unimodality}   & A         & 72.01     & 75.51 & 60.16     & 88.24 \\
                                & T         & 90.52     & 90.61 & 71.88     & 75.88\\
    \hline
                                &  A        & 48.76     & 86.53 & 60.51      & 84.12 \\
    Single Loss                    &  T        & 84.20     & 71.43 & 78.12     & 65.88 \\
                                &  A+T      & 89.62     & 88.57 & 73.44     & 81.18 \\
    \hline
                                & A         & 58.24     & 73.88 & 59.64     & 90.59 \\
    Perspective Loss                  & T         & 90.29     & 88.98 & 76.82     & 61.18 \\
                                & A+T       & 90.52     & 89.39 & 76.56     & 88.24\\
\bottomrule[1pt]
\end{tabular}}
\caption{The accuracy of unimodal model and multimodal model with ADD-Fusion at LGF on different emotions of IEMOCAP. A and T refer to audio and text modalities.}
\label{table:tab3}
\end{center}
\end{table}

\subsection{Single Loss v.s. Perspective Loss}
\label{singlevsperspective}
\noindent From Table~\ref{table:tab2}, we observe that multimodal models equipped with Perspective Loss bring obvious performance improvements in most fusion settings. To understand this point, we take LGF with ADD-Fusion as an example to study why Perspective Loss is so effective. Table~\ref{table:tab3} shows the detailed results of various modalities and models on different emotions.

For the unimodal structure, the text modality has excellent recognition ability in \texttt{happy} and \texttt{sad} emotions, while the audio modality performs well on \texttt{angry}. It is reasonable because users often use explicit wordings to express their \texttt{happy} and \texttt{sad} emotions, which makes text modality have better predictions on these two types. In contrast, it is easy to judge whether users are \texttt{angry} or not according to their tones, which are usually embodied in the audio signal.

For the multimodal model using Single Loss, the performance all decreases on the above three emotions. One possible reason is that text modality dominates information and audio modality is weakened when using simple fusion. Specifically, the multimodal model (A+T) has great performance degradation (up to 7\%) than the unimodal audio framework on \texttt{angry}. Besides, it also decreases slightly on \texttt{happy} and \texttt{sad} emotions because simple fusion mixes up multimodal features and obscures some key unimodal information.

After using Perspective Loss, the multimodal model can retain key discriminative unimodal characteristics of these specific emotions when performing fusion. We observe the model obtains the same or similar performance on \texttt{happy}, \texttt{sad}, and \texttt{angry} emotions. Additionally, it also achieves improvement in \texttt{neutral}. Finally, the two modalities play different roles and contribute to better overall performance.

\subsection{Proposed Method v.s. Previous Works}
\begin{table}[t]
\begin{center}
{
\begin{tabular}{l l}\toprule[1pt]
     \textbf{Approaches} & \textbf{WA(\%)} \\
    \hline
    bc-LSTM (2017)\cite{poria2017context}  & 75.60$^\dagger$\\
    CATF-LSTM (2017)\cite{poria2017multi}   & 80.10$^\dagger$    \\
    Zheng. (2019)\cite{lian2019conversational}      & 78.02 \\
    DANN (2020)\cite{lian2020context}     & 82.68    \\
    CONSK-GCN (2021)\cite{fu2021consk}               & 84.79$^{\ddagger}$ \\
    Soumya. (2022)\cite{dutta2022multimodal}        & 83.80 \\ \hline
    Our Method                                 & \textbf{85.40}$^*$    \\
\bottomrule[1pt]
\end{tabular}}
\caption{The performance of different methods on the IEMOCAP dataset. $^{\dagger}$ indicates results from audio+text modalities. $^{\ddagger}$ indicates the average result of the reproduction of open source code repeating 10 times. Our method is the multimodal model with LGF+ADD-Fusion+Perspective Loss. $^*$ represents our method outperforms other methods significantly ($p\text{-value}<0.01$).}
\label{tab4}
\end{center}
\end{table}

\noindent Table~\ref{tab4} shows the performance of different methods on the IEMOCAP dataset. By comparison, our method (LGF+ADD-Fusion+Perspective Loss) achieves a performance improvement of $0.61\%$ over the previous state-of-the-art method. This result indicates our method is simple but very effective for speech emotion recognition. 

Actually, previous works also use both audio and text modal features. However, they usually prematurely fuse multimodal features without perspective loss constraint, which may bear some noise information that is irrelevant to emotion. In contrast, LGF makes use of discriminative emotional information. Further, Perspective Loss makes each modality can play its own characteristics in predicting emotion. From Table~\ref{table:tab2} and Table~\ref{tab4}, we also see multimodal models without Perspective Loss are inferior to state-of-the-art methods. Perspective Loss facilitates emotion prediction significantly.

\section{Conclusion}
In this work, we empirically study speech emotion recognition from audio and text and explore the effect of different fusions of modalities. We find that the ADD-Fusion is a simple yet effective method. On this basis, we further propose Perspective Loss to retain the key and discriminative unimodal characteristics of specific emotions at fusion. The results show our method achieves state-of-the-art performance on the IEMOCAP dataset. We believe this study can be beneficial for other research on multimodal models and tasks.

\vfill\pagebreak

\bibliographystyle{IEEEbib}
\bibliography{refs}

\end{document}